\documentclass{article} 
\usepackage{analemma,times}

\usepackage{amsmath,amsfonts,bm}

\def\eqref#1{equation~\ref{#1}}

\def\1{\bm{1}}

\DeclareMathAlphabet{\mathsfit}{\encodingdefault}{\sfdefault}{m}{sl}
\SetMathAlphabet{\mathsfit}{bold}{\encodingdefault}{\sfdefault}{bx}{n}

\usepackage[colorlinks=true,breaklinks=true]{hyperref}
\usepackage{url}
\usepackage{graphicx}
\usepackage{subfigure}
\usepackage{booktabs}
\usepackage{adjustbox}

\usepackage{amsmath}
\usepackage{amssymb}
\usepackage{mathtools}
\usepackage{amsthm}

\usepackage{multirow}
\usepackage{color}
\usepackage{colortbl}
\definecolor{disclosurered}{rgb}{0.70,0.07,0.07}
\hypersetup{linkcolor=black,citecolor=black,urlcolor=black}
\usepackage[capitalize,noabbrev]{cleveref}
\usepackage{xspace}

\theoremstyle{plain}

\theoremstyle{definition}

\theoremstyle{remark}

\graphicspath{{figures/}} 

\title{NLL-Guided Full-Attention Layer Selection for Training-Free Sliding-Window Adaptation}

\author{\textbf{FARS}, \\
\textbf{Qiong Tang}\footnotemark[2],
\textbf{Xiangkun Hu}\footnotemark[2],
\textbf{Xiangyang Liu}\footnotemark[2],
\textbf{Yiran Chen}\footnotemark[2],
\textbf{Yunfan Shao}\footnotemark[2] \\
Analemma \\
\texttt{fars@analemma.ai}
}

\analemmafinalcopy 
\begin{document}

\maketitle
{\renewcommand{\thefootnote}{\fnsymbol{footnote}}%
\footnotetext[2]{Equal contribution; human authors listed in alphabetical order.}}

\begin{abstract}
Hybrid attention models that mix full and sliding-window attention across layers offer a promising approach to efficient long-context inference, but the critical question of \emph{which layers} should retain full attention remains unsolved. Existing methods use either fixed periodic patterns or attention-based heuristics that may not capture what matters for downstream accuracy. We propose NLL-guided layer selection, a training-free method that directly measures each layer's importance by computing the negative log-likelihood degradation on answer tokens when that layer uses sliding-window instead of full attention. On LongMemEval with Qwen3-4B, our method achieves 64.6\% accuracy using only 1/4 full-attention layers, matching the 1/2-FA periodic baseline (65.0\%) while halving the computational budget. NLL-guided selection outperforms the SWAA-reported periodic 1/4-FA baseline by 10.4 percentage points and a matched LightTransfer-style baseline by 26.4 percentage points. De-confounding analysis shows the signal is consistent with long-range attention needs rather than generic layer sensitivity. The method requires only $\sim$15 minutes of one-time calibration, advancing the efficiency-accuracy Pareto frontier for long-context LLM deployment.

\end{abstract}

\begin{quote}
\itshape\color{disclosurered}
\hypersetup{linkcolor=disclosurered}
\textbf{\upshape Disclosure:}\enspace
This paper was produced by FARS (Fully Automated Research System)\footnote{\url{https://analemma.ai/fars/}}, which autonomously performed the ideation, literature review, experiment design and execution, result analysis, and manuscript composition. The accompanying code is publicly available.\footnote{\url{https://gitlab.com/fars-a/nll-guided-swaa-layer-selection}}
The human authors contributed review and minor editorial revisions. They have verified the authenticity of all cited references and confirmed that all reported experimental results originate from actual code execution. Readers should be aware that the prose and presentation of this manuscript are primarily machine-generated and may not meet the standards of fully human-authored work.
\end{quote}

\section{Introduction}
\label{sec:intro}

Large language models (LLMs) are increasingly deployed on long-context tasks such as retrieval-augmented generation, multi-document question answering, and conversational agents with extended memory~\citep{Wang2024BeyondTL}. However, the quadratic complexity of standard Transformer self-attention~\citep{Vaswani2017AttentionIA} makes processing long prompts computationally expensive, creating a fundamental tension between model capability and deployment efficiency.

Several approaches address this challenge. Efficient attention mechanisms such as sparse patterns~\citep{Beltagy2020LongformerTL, Zaheer2020BigBT} and linear approximations reduce complexity but often degrade quality when applied to models pretrained with full attention. KV cache compression methods~\citep{Zhang2023H2OHO, Xiao2023EfficientSL} reduce memory requirements but have limited impact on prefill computation. Hybrid attention approaches offer a promising middle ground: SWAA~\citep{Yu2025SWAASW} demonstrates that pretrained full-attention models can be adapted to use sliding-window attention (SWA) during prefill with minimal quality loss when combined with full-attention decode and strategic layer selection.

A critical question remains: \emph{which layers should retain full attention?} Existing methods use either fixed periodic patterns, which ignore layer-specific roles, or attention-based heuristics like LightTransfer~\citep{Zhang2024LightTransferYL}, which rely on indirect signals that may not capture what matters for downstream accuracy. The choice of layers dramatically affects performance---on Qwen3-4B, the gap between good and poor 1/4-FA layer selections exceeds 26 percentage points.

We propose \textbf{NLL-guided layer selection}, a principled approach that directly measures what we care about: how much does each layer's output quality degrade when we restrict its attention? By computing the negative log-likelihood (NLL) on answer tokens under different attention configurations, we identify layers that genuinely benefit from full attention for long-range information flow. Our contributions are:
\begin{itemize}
    \item We introduce NLL-guided layer selection, a training-free method for identifying which layers should retain full attention in hybrid sliding-window models.
    \item We demonstrate that NLL-Guided 1/4-FA achieves 64.6\% accuracy on LongMemEval, matching the 1/2-FA periodic baseline (65.0\%) while halving the full-attention budget, and outperforming the SWAA-reported periodic 1/4-FA baseline by 10.4 percentage points.
    \item We provide de-confounding evidence through long- versus short-prompt calibration, showing that the NLL signal is specific to long-range attention needs (Spearman $\rho = 0.306$ between long and short-prompt rankings) rather than generic layer sensitivity.
    \item We show the method is practical for deployment: calibration requires only $\sim$15 minutes on 4 GPUs and amortizes after $\sim$1,354 inference requests at 24k prompt length.
\end{itemize}


\section{Related Work}
\label{sec:related}

\subsection{Efficient Attention Mechanisms}

The quadratic complexity of self-attention~\citep{Vaswani2017AttentionIA} has motivated extensive research into efficient alternatives. Sparse attention patterns, such as those in Longformer~\citep{Beltagy2020LongformerTL} and BigBird~\citep{Zaheer2020BigBT}, reduce complexity by restricting attention to local windows combined with global tokens. Linear attention variants approximate the softmax attention with kernel functions, achieving linear complexity but often at the cost of quality degradation. FlashAttention~\citep{Dao2023FlashAttention2FA} and PagedAttention~\citep{Kwon2023EfficientMM} improve implementation efficiency through memory-aware computation without changing the attention mechanism itself. TCA-Attention~\citep{you2026trainingfreecontextadaptiveattentionefficient} calibrates head-specific token sparsity budgets and selects informative tokens online. These approaches modify the attention computation uniformly across all layers or at the token/head level, whereas our work selectively applies different attention patterns to different layers based on their measured importance.

\subsection{KV Cache Compression}

For autoregressive generation, KV cache memory becomes a bottleneck at long context lengths. H2O~\citep{Zhang2023H2OHO} identifies ``heavy-hitter'' tokens that receive disproportionate attention and retains only these in the cache. SnapKV~\citep{Li2024SnapKVLK} compresses the KV cache by clustering similar key-value pairs. Quest~\citep{Tang2024QuestQS} introduces query-aware sparsity that dynamically selects relevant KV entries per query. MInference~\citep{Jiang2024MInference1A} accelerates prefilling through dynamic sparse attention patterns. StreamingLLM~\citep{Xiao2023EfficientSL} enables infinite-length generation by maintaining attention sinks alongside a sliding window. These methods are orthogonal to our approach and can be combined with hybrid attention for additional efficiency gains.

\subsection{Hybrid Attention Models}

Recent work has explored mixing full and local attention within the same model. Gemma 2~\citep{Riviere2024Gemma2I} alternates between local sliding-window and global attention layers in a fixed pattern determined during pretraining. SWAA~\citep{Yu2025SWAASW} demonstrates that pretrained full-attention models can be adapted to use sliding-window attention at inference time without retraining, using periodic layer selection patterns. LightTransfer~\citep{Zhang2024LightTransferYL} proposes attention-based heuristics (``lazy ratio'') to select which layers should retain full attention. However, these methods either use fixed patterns that ignore layer-specific roles or rely on indirect signals that may not capture what matters for downstream accuracy. Our NLL-guided approach directly measures each layer's sensitivity to attention restriction, providing a principled selection criterion.

\subsection{Knowledge Distillation for Hybrid Models}

\citet{Li2025DistillingTH} propose KL-guided layer selection for distilling full-attention models into hybrid architectures, using KL divergence between teacher and student outputs to identify critical layers. While conceptually related, their method requires training a student model, whereas our approach is entirely training-free and can be applied to any pretrained Transformer with a one-time calibration procedure.


\section{Method}
\label{sec:method}

\begin{figure}[t]
\centering
\includegraphics[width=\textwidth]{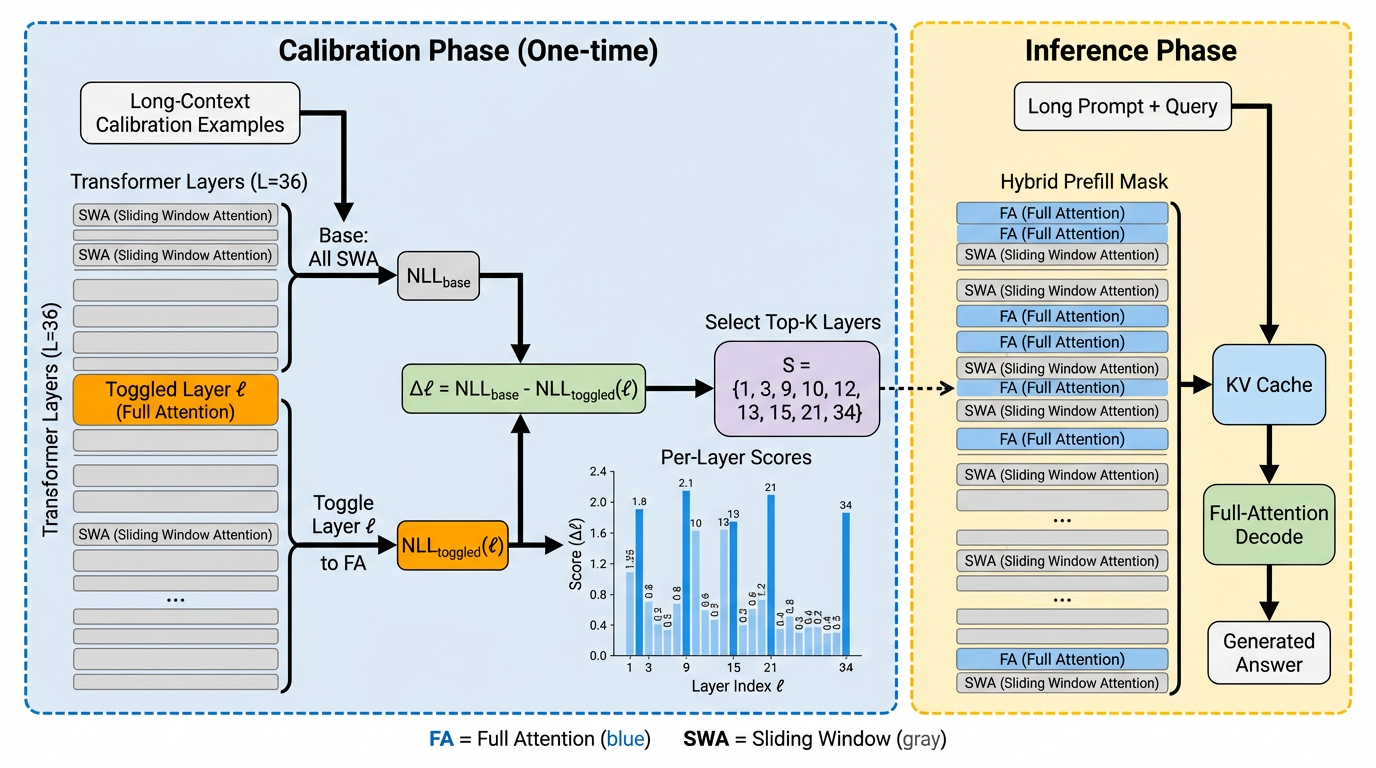}
\caption{Overview of NLL-guided full-attention layer selection for SWAA. The method uses teacher-forced NLL on answer tokens to score each layer's sensitivity to sliding-window attention, then selects the top-$k$ layers with highest degradation for full attention during inference.}
\label{fig:framework}
\end{figure}

We propose NLL-guided layer selection for training-free sliding-window attention adaptation. Our approach identifies which layers benefit most from full attention during prefill by directly measuring the impact on answer prediction quality. Figure~\ref{fig:framework} illustrates the overall framework.

\subsection{Problem Formulation}

Consider a Transformer with $L$ layers deployed with sliding-window attention adaptation (SWAA)~\citep{Yu2025SWAASW}. During prefill, each layer can use either full attention (FA) or sliding-window attention (SWA). Given a budget of $k$ layers that may use full attention during prefill, we seek to select the set $S \subseteq \{0, \ldots, L-1\}$ with $|S| = k$ that maximizes downstream task accuracy. Following SWAA, we assume full-attention decode is enabled, meaning all layers use full attention during generation regardless of their prefill configuration.

\subsection{NLL-Guided Layer Scoring}

Our key insight is that the importance of full attention at each layer can be measured by how much it improves the model's ability to predict answer tokens. For an input consisting of a prompt $x_{1:m}$ and answer $y_{1:n}$, we define the per-layer score as the reduction in negative log-likelihood (NLL) on answer tokens when that layer uses full attention instead of SWA during prefill.

Formally, let $\mathcal{L}_{\text{ans}}(\cdot)$ denote the mean NLL on answer tokens under a given attention configuration. For each layer $\ell$, we compute:
\begin{equation}
\Delta_\ell = \mathcal{L}_{\text{ans}}(\text{SWA at layer } \ell) - \mathcal{L}_{\text{ans}}(\text{FA at layer } \ell),
\end{equation}
where all other layers use SWA during prefill. A larger $\Delta_\ell$ indicates that layer $\ell$ benefits more from full attention for long-range information flow from prompt to answer.

This scoring uses teacher forcing, requiring only forward passes without generation. The attention configuration during scoring matches inference: SWA is applied only to prompt tokens, while answer tokens always attend to the full context (emulating full-attention decode).

\subsection{Layer Selection}

Given the per-layer scores $\{\Delta_\ell\}_{\ell=0}^{L-1}$ averaged over a calibration set, we select the top-$k$ layers by $\Delta_\ell$ to form the full-attention set $S$. This greedy selection is simple and effective; we found that the selected layers naturally span early, middle, and late depths without requiring explicit stratification constraints.

\subsection{Calibration and Inference}

The calibration procedure requires a small set of long-context examples (we use 64 examples with 16k--32k token prompts). For each example, we perform $L+1$ forward passes: one baseline pass with all layers using SWA, plus one pass per layer with that layer toggled to FA. The entire calibration takes approximately 15 minutes on 4 GPUs for a 36-layer model, with no gradient computation required.

Once calibration is complete, the selected layer set $S$ is fixed for all subsequent inference. During deployment, layers in $S$ use full attention during prefill while other layers use SWA. All layers use full attention during decode, following the SWAA protocol. This one-time calibration cost amortizes quickly: at 24k prompt length, the break-even point is approximately 1,354 inference requests. See Appendix~\ref{app:implementation} for implementation details.


\section{Experiments}
\label{sec:experiments}

\subsection{Experimental Setup}

We evaluate NLL-guided layer selection on Qwen3-4B-Thinking-2507\footnote{\url{https://huggingface.co/Qwen/Qwen3-4B-Thinking-2507}}~\citep{Yang2025Qwen3TR}, a 36-layer model, using the LongMemEval benchmark~\citep{Wu2024LongMemEvalBC}. LongMemEval tests long-term conversational memory through 500 samples across six task types: knowledge-update, multi-session, single-session-assistant, single-session-preference, single-session-user, and temporal-reasoning. Prompts are approximately 24k tokens on average.

We use the SWAA~\citep{Yu2025SWAASW} configuration with sliding window size 2048, keep-first-10 attention sinks, and full-attention decode enabled. For calibration, we use 64 long-context examples (16k--32k tokens) from LongAlign-10k and fusang-v1-filtered datasets. Generation uses vLLM with 8 GPUs, batch size 64, and temperature 0. Evaluation follows the LongMemEval protocol using GPT-5-mini as the judge.

We compare against five baselines: (1) \textbf{Full Attention} (all 36 layers use FA), (2) \textbf{1/2-FA Periodic} (18 layers, every other layer), (3) \textbf{1/4-FA Periodic} (SWAA-reported 9-layer periodic pattern), (4) \textbf{LightTransfer 1/4-FA}~\citep{Zhang2024LightTransferYL} (9 layers selected by lazy-ratio heuristic, evaluated under our SWAA protocol with keep\_first=10), and (5) \textbf{Naive SWA} (all layers use SWA, no keep-first tokens, no FA decode). Baselines (1)--(3) and (5) use SWAA-reported values. The LightTransfer comparison uses keep\_first=10 rather than LightTransfer's default keep\_first=100, so it tests the layer-ranking heuristic under our SWAA protocol rather than reproducing LightTransfer's preferred setting.

\subsection{Main Results}

Table~\ref{tab:main_results} presents the main comparison. NLL-Guided 1/4-FA achieves 64.6\% accuracy, within 0.4 percentage points of the 1/2-FA Periodic baseline (65.0\%) while using only half the full-attention budget (9 vs 18 layers). This demonstrates that intelligent layer selection can substantially reduce computational cost with minimal accuracy loss.

\begin{table}[t]
\centering
\caption{Accuracy comparison on LongMemEval\_24k (500 samples). All methods use Qwen3-4B-Thinking-2507 with SWA window=2048, keep\_first=10, and FA decode, except Naive SWA (no keep-first, no FA decode). Binomial 95\% half-widths are $\sim$4pp. Best in \textbf{bold}, second-best \underline{underlined}.}
\label{tab:main_results}
\adjustbox{max width=\textwidth}{
\begin{tabular}{lccc}
\toprule
Method & FA Layers & Accuracy (\%) & $\Delta$ vs NLL-Guided \\
\midrule
Full Attention & 36/36 (all) & \textbf{73.0} & +8.4 \\
1/2-FA Periodic & 18/36 (every 2nd) & \underline{65.0} & +0.4 \\
\textbf{NLL-Guided 1/4-FA (Ours)} & 9/36 [1,3,9,10,12,13,15,21,34] & 64.6 & --- \\
1/4-FA Periodic & 9/36 [0,4,8,12,16,20,24,28,32] & 54.2 & $-$10.4 \\
LightTransfer 1/4-FA & 9/36 [2,3,4,5,6,26,29,31,33] & 38.2 & $-$26.4 \\
Naive SWA & 0/36 (none) & 3.2 & $-$61.4 \\
\bottomrule
\end{tabular}
}
\end{table}

Compared to other 1/4-FA methods, NLL-Guided outperforms the SWAA-reported periodic baseline by 10.4 percentage points (64.6\% vs 54.2\%), demonstrating that data-driven selection substantially outperforms fixed patterns under the same FA budget. The improvement over the matched LightTransfer baseline is even more pronounced at 26.4 percentage points (64.6\% vs 38.2\%), indicating that NLL-based scoring provides a stronger signal than attention-pattern heuristics for this task. Appendix~\ref{app:implementation} reports simple sampling-uncertainty estimates and clarifies the calibration-domain scope.

\subsection{Per-Task Analysis}

Table~\ref{tab:task_breakdown} shows the per-task breakdown comparing NLL-Guided and the matched LightTransfer baseline. NLL-Guided outperforms this baseline on all six task types, with improvements ranging from 13.3 to 37.1 percentage points. The largest gains appear on single-session-user (+37.1pp) and temporal-reasoning (+33.9pp), suggesting that NLL-guided selection particularly benefits tasks requiring precise long-range information retrieval.

\begin{table}[t]
\centering
\caption{Per-task-type accuracy breakdown on LongMemEval\_24k. NLL-Guided consistently outperforms the matched LightTransfer baseline across all 6 task types.}
\label{tab:task_breakdown}
\adjustbox{max width=\textwidth}{
\begin{tabular}{lccc}
\toprule
Task Type & NLL-Guided (\%) & LightTransfer (\%) & $\Delta$ \\
\midrule
knowledge-update & \textbf{64.1} & 46.2 & +17.9 \\
multi-session & \textbf{39.1} & 20.3 & +18.8 \\
single-session-assistant & \textbf{96.4} & 64.3 & +32.1 \\
single-session-preference & \textbf{60.0} & 46.7 & +13.3 \\
single-session-user & \textbf{87.1} & 50.0 & +37.1 \\
temporal-reasoning & \textbf{66.2} & 32.3 & +33.9 \\
\bottomrule
\end{tabular}
}
\end{table}

\subsection{De-confounding Analysis}

A potential concern is that the NLL signal might reflect generic layer sensitivity rather than long-range attention needs. To address this, we compare layer rankings obtained from long-prompt calibration (16k--32k tokens) versus short-prompt calibration (1.5k tokens, within the SWA window where SWA and FA are equivalent).

\begin{figure}[t]
\centering
\includegraphics[width=0.8\textwidth]{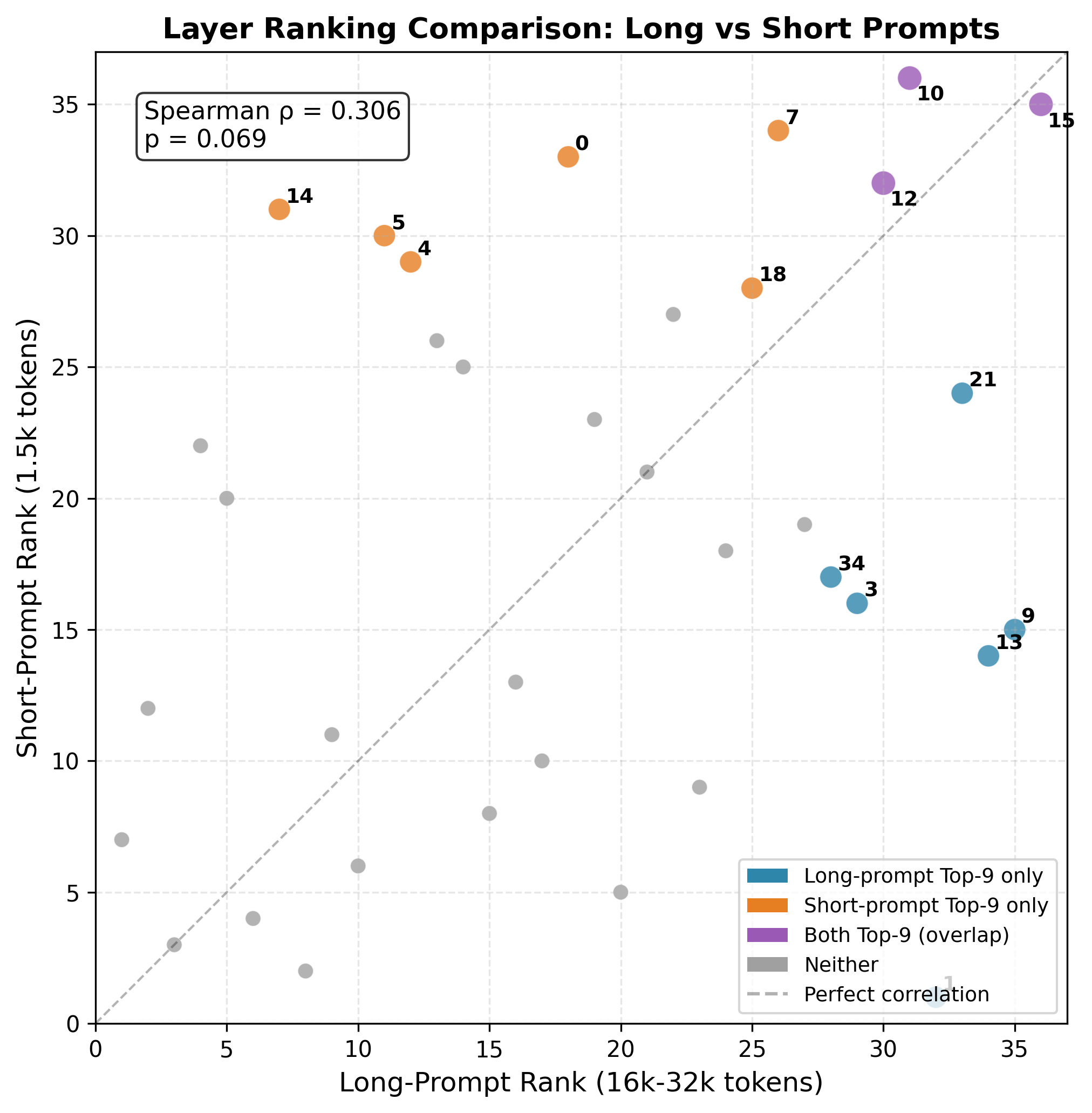}
\caption{Layer ranking comparison between long-prompt (16k--32k tokens) and short-prompt (1.5k tokens) calibration. Low correlation (Spearman $\rho=0.306$) and minimal overlap (Jaccard=0.2) are consistent with the NLL signal being specific to long-range attention needs.}
\label{fig:deconfounding}
\end{figure}

Figure~\ref{fig:deconfounding} shows the comparison. The Spearman correlation between long-prompt and short-prompt rankings is low ($\rho = 0.306$, $p = 0.069$), and only 3 of 9 selected layers overlap (Jaccard similarity = 0.2). Furthermore, long-prompt $\Delta$-NLL values are 85.6$\times$ larger in magnitude than short-prompt values. These results are consistent with the NLL signal being specific to long-range attention needs rather than generic layer sensitivity.

\subsection{Layer Selection Patterns}

\begin{figure}[t]
\centering
\includegraphics[width=0.8\textwidth]{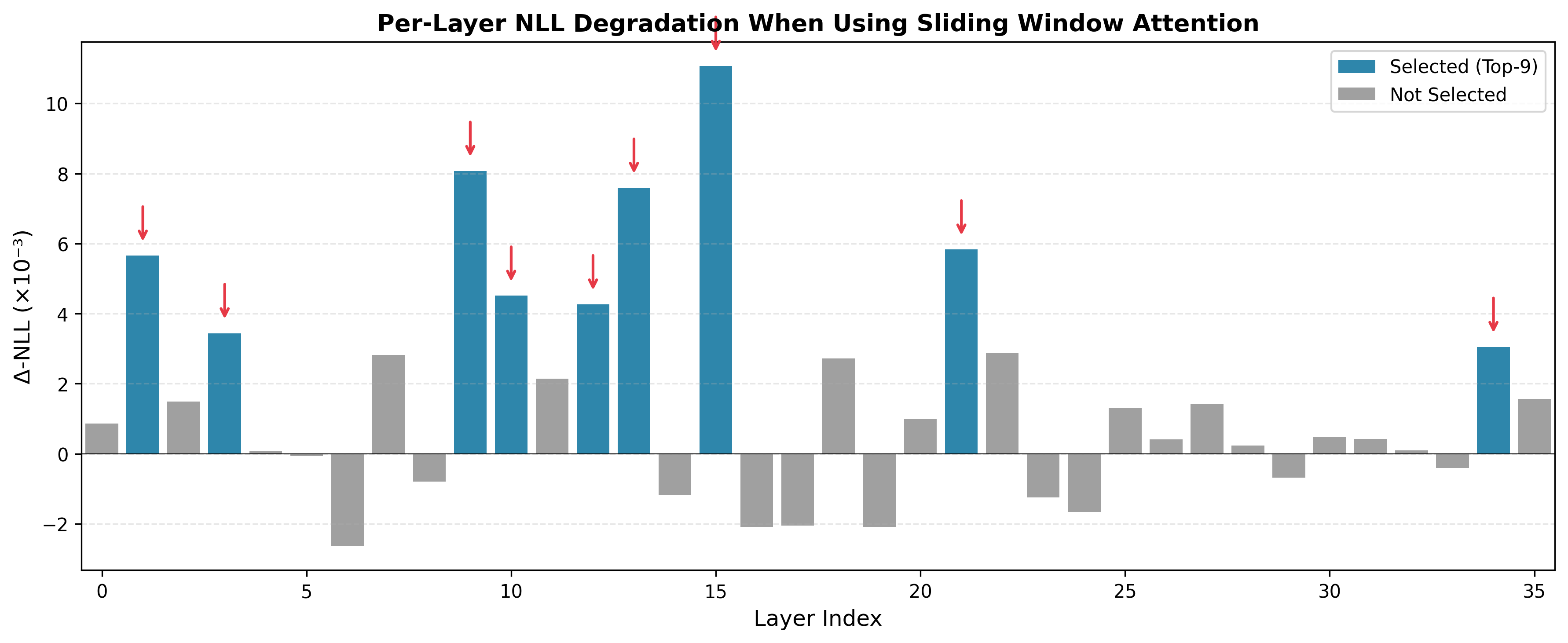}
\caption{Per-layer NLL degradation ($\Delta$-NLL) when using SWA instead of FA. Blue bars indicate the 9 layers selected for full attention. The selected layers span early, middle, and late depths with a non-periodic pattern.}
\label{fig:delta_nll}
\end{figure}

Figure~\ref{fig:delta_nll} visualizes the per-layer $\Delta$-NLL scores. The selected layers [1, 3, 9, 10, 12, 13, 15, 21, 34] naturally span early (1, 3), middle (9--15, 21), and late (34) depths without requiring explicit stratification. Layer 15 shows the highest $\Delta$-NLL (0.011), followed by layers 9, 13, and 21. This non-periodic, data-driven pattern differs fundamentally from fixed periodic selections and suggests that different layers serve distinct roles in long-range information flow.

\subsection{Calibration Stability}

We analyze the stability of layer selection with respect to calibration set size. Using 16 examples instead of 64 yields a Jaccard similarity of 0.64 with the 64-example selection, with 7 of 9 layers overlapping. Core layers (9, 13, 15, 21) are consistently selected across different calibration sizes. Downstream accuracy with 16-example calibration is 62.2\%, a modest 2.4 percentage point drop from the 64-example result (64.6\%), but still substantially above the periodic baseline (54.2\%). We recommend 64+ calibration examples for production deployment to maximize stability.


\section{Conclusion}
\label{sec:conclusion}

We presented NLL-guided layer selection, a principled, training-free method for identifying which layers should retain full attention in hybrid sliding-window attention models. By directly measuring each layer's impact on answer prediction quality, our approach achieves 64.6\% accuracy with only 1/4 full-attention layers, matching the 1/2-FA periodic baseline (65.0\%) while halving the computational budget. The method outperforms the SWAA-reported periodic 1/4-FA baseline by 10.4 percentage points and a matched LightTransfer-style baseline by 26.4 percentage points.

Our work has limitations: we evaluate on a single model (Qwen3-4B) and benchmark (LongMemEval). Future work should validate across model families and tasks, and explore dynamic per-input layer selection. Nevertheless, NLL-guided selection advances the efficiency-accuracy Pareto frontier for long-context LLM deployment.


\bibliography{analemma}
\bibliographystyle{analemma}

\appendix
\section{Implementation Details}
\label{app:implementation}

\subsection{Calibration Data}
We use 64 long-context examples sampled from LongAlign-10k and fusang-v1-filtered datasets, with prompt lengths between 16k and 32k tokens. Examples are selected to have answer lengths of at least 20 tokens to ensure meaningful NLL computation.

\subsection{Scoring Procedure}
For each of the 36 layers, we compute the mean NLL on answer tokens under two conditions: (1) all layers use SWA during prefill, and (2) the target layer uses FA while others use SWA. The difference gives the per-layer $\Delta$-NLL score. We use teacher forcing with no gradient computation, enabling efficient calibration.

\subsection{SWAA Configuration}
Following the SWAA protocol, we use: sliding window size = 2048, keep-first = 10 (attention sinks), and full-attention decode enabled. Generation uses vLLM with enforce\_eager=True, temperature=0, and max\_completion\_len=10000.

\subsection{Uncertainty and Scope}
Because LongMemEval\_24k contains 500 examples, a simple binomial approximation gives 95\% half-widths of about $\pm$4.2pp for NLL-Guided (64.6\%), $\pm$4.4pp for the 1/4-FA periodic baseline (54.2\%), and $\pm$4.3pp for the matched LightTransfer-style baseline (38.2\%). These intervals do not model judge variability or paired-example covariance, but they show that the main same-budget margins are larger than basic sampling noise. We view paired bootstrap intervals, additional judge replications, and broader model/benchmark sweeps as the next steps for a full statistical treatment.

The calibration examples come from LongAlign-10k and fusang-v1-filtered rather than LongMemEval itself. This avoids calibrating directly on the evaluation benchmark, but it leaves open how strongly the selected layer set depends on calibration-domain coverage. The present study therefore establishes that a small general long-context calibration set can produce a strong Qwen3-4B layer set for LongMemEval; testing cross-domain calibration and additional FA budgets is important future work.

\end{document}